\documentclass[10pt,twocolumn,letterpaper]{article}

\usepackage{cvpr}              
\usepackage[accsupp]{axessibility}

\definecolor{cvprblue}{rgb}{0.21,0.49,0.74}
\usepackage[pagebackref,breaklinks,colorlinks,allcolors=cvprblue]{hyperref}

\def\paperID{560} 
\def\confName{CVPR}
\def\confYear{2025}

\title{Blind Bitstream-corrupted Video Recovery via Metadata-guided Diffusion Model}

\author{
    Shuyun Wang$^{1,2}$ \quad 
    Hu Zhang$^2$ \quad 
    Xin Shen$^1$ \quad
    Dadong Wang$^2$ \quad
    Xin Yu$^1$\thanks{Corresponding author}\\
    $^1$The University of Queensland \quad 
    $^2$Data61, CSIRO, Australia \\
    {\tt\small \{shuyun.wang, xin.yu\}@uq.edu.au}
}

\begin{document}
\twocolumn[{
\maketitle
\vspace{-2em}
\begin{center}
    \centering
    \includegraphics[width=\textwidth]{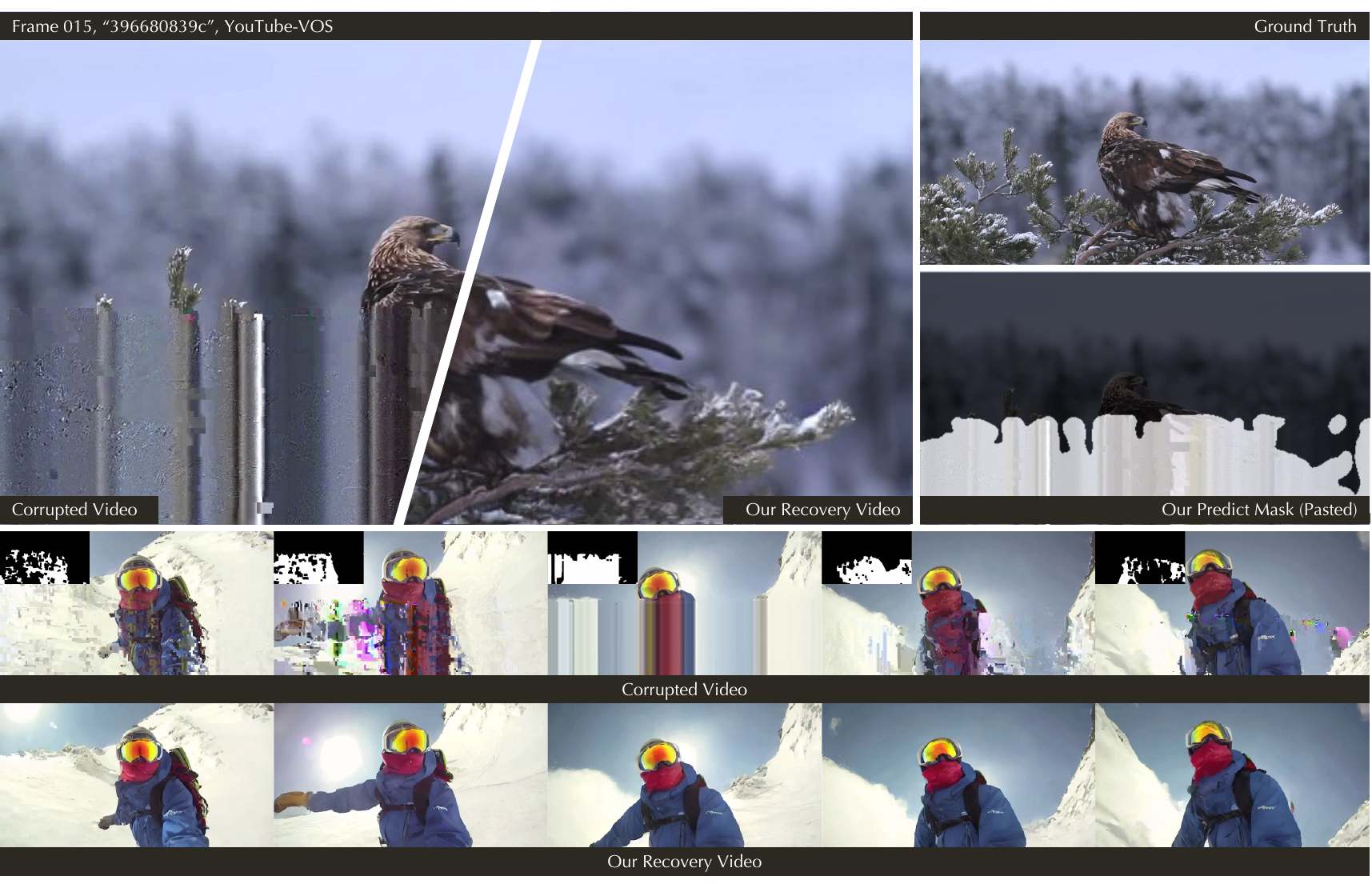}
    \vspace{-1em}
    \captionof{figure}{\textbf{Blind Bitstream-corrupted Video Recovery.} Our method achieves superior recovery without giving any predefined masks during inference and precisely predicts the location of corrupted regions.}
    \label{fig:1}
\end{center}
}]
\setcounter{footnote}{1}
\footnotetext[1]{Corresponding author}

\begin{abstract}
Bitstream-corrupted video recovery aims to fill in realistic video content due to bitstream corruption during video storage or transmission.
Most existing methods typically assume that the predefined masks of the corrupted regions are known in advance.
However, manually annotating these masks is laborious and time-consuming, limiting the applicability of existing methods in real-world scenarios. 
Therefore, we expect to relax this assumption by defining a new blind video recovery setting where the recovery of corrupted regions does not rely on predefined masks.
There are two significant challenges in this setting: 
(i) without predefined masks, how accurately can a model identify the regions requiring recovery?
(ii) how to recover contents from extensive and irregular regions, especially when large portions of frames are severely degraded?
To address these challenges, we introduce a Metadata-Guided Diffusion Model, dubbed M-GDM.
To enable a diffusion model focusing on the corrupted regions, we leverage intrinsic video metadata as a corruption indicator and design a dual-stream metadata encoder.
This encoder first embeds the motion vectors and frame types of a video separately and then merges them into a unified metadata representation.
The metadata representation will interact with the corrupted latent feature through cross-attention mechanisms at each diffusion step.
Meanwhile, to preserve the intact regions, we propose a prior-driven mask predictor that generates pseudo masks for the corrupted regions by leveraging the metadata prior and diffusion prior.
These pseudo masks enable the separation and recombination of intact and recovered regions through hard masking. 
However, imperfections in pseudo mask predictions and hard masking processes often result in boundary artifacts. 
Thus, we introduce a post-refinement module that refines the hard-masked outputs, enhancing the consistency between intact and recovered regions.
Extensive experiment results validate the effectiveness of our method and demonstrate its superiority in the blind video recovery task.
Our code is available at \url{https://github.com/Shuyun-Wang/M-GDM}.

\end{abstract} 
\section{Introduction}
\label{sec:intro}

Bitstream-corrupted video recovery~\cite{liu2024bitstream} aims to restore visually coherent content in videos that have been degraded due to ``bit loss'' during compression, storage, and transmission. 
This task is essential due to its broader applications, like digital archiving.
Meanwhile, it is quite challenging since those corruptions often introduce unpredictable and irreversible errors into decoded frames, leading to serious and complex corruption patterns.
For effective recovery, the corrupted areas should be smoothly blended with the surroundings and filled with natural content, and the original intact regions should be preserved.

\begin{figure}[t]
  \centering
   \includegraphics[width=0.8\linewidth]{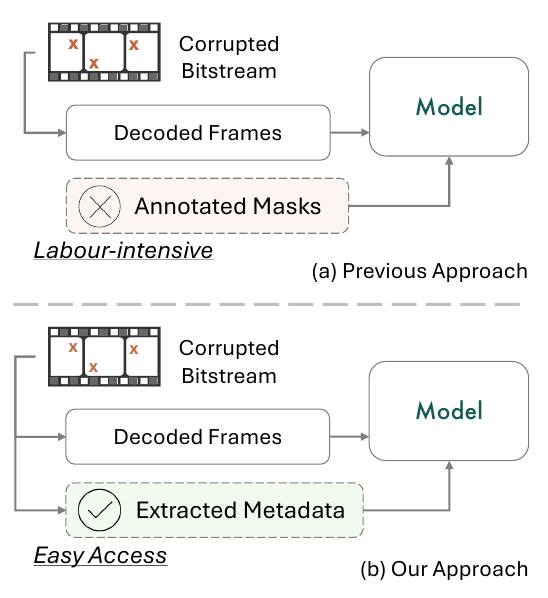}
   \vspace{-1em}
    \caption{\textbf{Comparison of the Previous and Proposed Approaches.} (a) Previous approaches rely on predefined input masks, which are labor-intensive and difficult to obtain. (b) Our approach employs easily accessible metadata extracted from the bitstream as guidance without using any predefined masks. }
   \label{fig:2}
\end{figure}

Most existing methods~\cite{zhang2023avid, Zheng_2023_ICCV, wu2023semi, zhou2023propainter, li2022towards} have a strong assumption that accurately annotated masks of the corrupted areas are available, as shown in Fig.~\ref{fig:2} (a), which means the regions to be recovered are known in advance.
However, such an assumption is overly idealistic and impractical in real-world scenarios.
A video of just a few seconds may contain hundreds of millions of pixels, making it laborious and infeasible to manually annotate which pixels are corrupted. 
As a result, these methods face significant limitations when trying to apply them to practical applications.

In this paper, we consider a new blind bitstream-corrupted video recovery setting, where the model is required to restore the corrupted regions in a video without any predefined masks.
To address this challenge, we propose a Metadata-Guided Diffusion Model (M-GDM), specifically designed for this novel problem.
M-GDM builds upon a latent diffusion model~\cite{ldm}, leveraging the powerful generative capabilities of the diffusion model to handle severe, large-scale corruptions.
To guide the diffusion model concentrated on the corrupted regions, we leverage inherent video metadata, as shown in Fig.~\ref{fig:2} (b), as a corruption indicator and design a dual-stream metadata encoder.
We find the video metadata, including motion vectors and frame types, strongly correlates with corruption patterns.
Thus, they can be used to indicate the corrupted regions. 
The metadata encoder is composed of two modality-specific encoders to deal with motion vectors and frame types separately.
We extract the corresponding embeddings and aggregate them into the latent space.
Throughout the diffusion process, the aggregated representation will be injected into the cross-attention layers to interact with the corrupted frame latent. 
The metadata can help us guide the diffusion model to focus on the corrupted areas to better exploit the generative prior.

To preserve the intact area from being modified, we introduce a prior-driven mask predictor generating pseudo masks for corrupted regions. 
The predictor receives attention results from the diffusion model and metadata representations from the dual-stream metadata encoder, which stands for the diffusion prior and the metadata prior, respectively.
To supervise the pseudo masks in training, we use binarized residuals between corrupted frames and ground-truth frames as supervision. 
The estimated pseudo masks play the same role as previous predefined masks.
They will be used by the hard masking mechanism to separate intact regions from corrupted frames and recovered regions from the diffusion model output.
Such a process will eliminate the potential negative effects on intact regions.
However, pseudo masks are not always perfect, and there will exist some gradients at the region boundaries due to the hard masking mechanism.
For this purpose, we design a post-refinement module that contains stacked residual Swin transformer blocks to keep the content consistent.

Together with all the proposed components, our approach achieves visually appealing recovery results and eliminates the need for pre-defined masks.
As shown in Fig.~\ref{fig:1}, we can generate realistic eagle details and plant textures.
In addition, the skiing scenario in Fig.~\ref{fig:1} also indicates that we can recover consistent video content even under complex corruption patterns.
Thanks to our special design and pretrained generative prior, our model achieves state-of-the-art performance on existing benchmarks, showing remarkable visual realism.
In summary, our key contributions are as follows:
\begin{itemize} 
\item Our work is the first attempt to investigate the blind bitstream-corrupted video recovery task. It eliminates the need for predefined masks and is more desirable for real-world applications.
\item We propose a metadata-guided diffusion model (M-GDM) that consists of three novel components. It predicts corrupted regions and restores them within a unified framework, making blind video recovery feasible.
\item M-GDM outperforms baseline methods in both qualitative and quantitative performance. It effectively recovers corrupted areas with temporal and spatial consistent visual details without requiring predefined masks.
\end{itemize}
\section{Related Work}
\label{sec:Relatedwork}
\subsection{Traditional Video Recovery}
Traditional video recovery typically focuses on simulated and manual-designed missing areas within the decoded video content.
The well-known existing methods include video inpainting, completion, and error concealment. Video inpainting is similar to video completion, which aims
to complete missing regions in a given video with predefined corruption masks~\cite{xu2019deep}. In video inpainting/completion, flow-guided generative methods demonstrate strong recovery capabilities by utilizing optical flow to capture spatial and temporal relationships between frames~\cite{zeng2020learning, gao2020flow, lao2021flow, kang2022error, zhang2022inertia, Zheng_2023_ICCV,zhang2023avid}. Xu~\etal~\cite{xu2019deep} focuses on first completing the flow and then using it as guidance for pixel-domain propagation. 
Li{~\etal}~\cite{li2022towards} proposes an end-to-end video inpainting framework that jointly learns flow completion and feature propagation in the down-sampled feature domain. Zhou{~\etal}~\cite{zhou2023propainter} introduces a dual-domain propagation approach, combining global image and local feature propagation with a mask-guided sparse video Transformer to achieve efficient and high-quality video inpainting. Recently, Wu{~\etal}~\cite{wu2024towards} presents language-driven video inpainting, replacing manual masks with natural language guidance and using a diffusion-based model to handle interactive and referring inpainting tasks.

Video error concealment, on the other hand, focuses more on the post-processing at the decoder side~\cite{wu2023spatial}. Traditional methods include improvements on spatial, temporal, and hybrid spatialtemporal~\cite{ye2008hybrid, koloda2013sequential, chang2013motion, lin2013error, kazemi2021review}. Recently, some deep learning-based methods are proposed~\cite{wu2023spatial, xiang2019generative, chung2019bi}. For example, in~\cite{sankisa2018video}, they adopt an adaptable decoder-like model for video error concealment through optical flow prediction using deep neural networks.

\subsection{Bitstream-corrupted Video Recovery}
Different from traditional video recovery tasks that address manual-designed error areas with slice or block shapes, bitstream-corrupted video recovery targets bitstream errors in real-world multimedia communications. This task aims to restore visually coherent content that seamlessly integrates with surrounding areas from genuinely corrupted bitstreams. Recently, Liu{~\etal}~\cite{liu2024bitstream} introduce the first large-scale dataset specifically for bitstream-corrupted video recovery, featuring realistic corruption patterns.
Nevertheless, their solution is still based on traditional methods~\cite{zeng2020learning, liu2021fuseformer} and rely on predefined masks to locate the corrupted regions during inference. The usage of these masks is impractical since real-world corruptions are generally unpredictable and irregular. Besides, annotating these masks is labor-intensive and costly, limiting the applicability of these methods in real-world scenarios. 

In this work, we propose the first blind bitstream-corrupted video recovery that eliminates the need for predefined masks, enabling a more realistic and adaptable solution for real-world applications. Inspired by recent efforts of exploring strong generation prior for video editing~\cite{ceylan2023pix2video,geyer2023tokenflow,qi2023fatezero,wu2023tune,wang2024videocomposer}, we base our method on diffusion models~\cite{song2020denoising,ldm} to address the challenges in blind bitstream-corrupted video recovery.

\section{Methodology}
\label{sec:method}
\begin{figure*}[t]
  \centering
  \vspace{-1em}
   \includegraphics[width=0.95\linewidth]{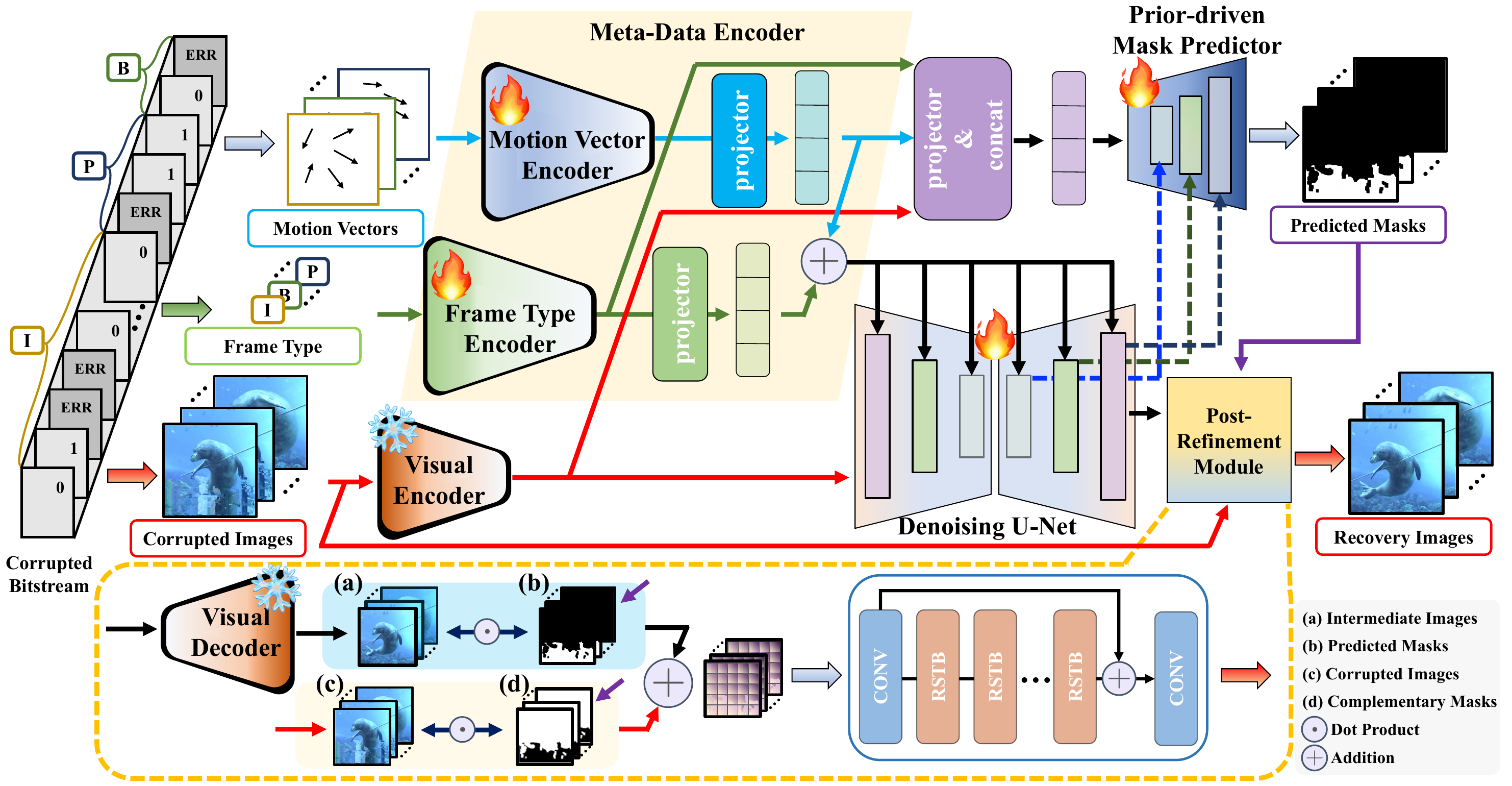}
    \caption{\textbf{Overview of Our Proposed Metadata-Guided Diffusion Model (M-GDM).} M-GDM consists of three key components for corrupted bitstream video recovery: (1) a dual-stream metadata encoder that extracts metadata representations from the corrupted bitstream, (2) a prior-driven mask predictor that predicts masks for corrupted regions, and (3) a post-refinement module that eliminates boundary artifacts after hard-masking, where RSTB represents a residual swin transformer block. The metadata guides the diffusion model to focus on corrupted areas, while the prior-driven mask predictor and post-refinement module facilitate a better fusion of intact and recovered regions, resulting in high quality video recovery. `0' and `1' represent the raw bitstream.}
   \label{fig:3}
\end{figure*}
\noindent
{\bf Notations:}
Let $X=\{{x_i\in \mathbb{R}^{H\times W\times 3}}\}^N_{i=1}$ represent the decoded corrupted frames from the video bitstream, where $N$ denotes the number of frames. Our objective is to restore the video frames $\hat{Y}=\{{\hat{y}_i\in \mathbb{R}^{H\times W\times 3}}\}^N_{i=1}$. $Y=\{{y_i\in \mathbb{R}^{H\times W\times 3}}\}^N_{i=1}$ represents the corresponding ground truth frames for comparison. We expect $\hat{Y}$ to be closely aligned with $Y$ in both spatial and temporal dimensions. The motion vectors between frames are denoted as $V=\{{{v}_i\in \mathbb{R}^{H\times W\times 4}}\}^N_{i=1}$ and the frame type of each frame is denoted as $F=\{{{f}_i}\in \mathbb{R}^{N}\}^N_{i=1}$. For simplicity, we use one frame as an example.

\noindent
{\bf Preliminaries:} We build our framework on a diffusion-based model~\cite{ldm}. The diffusion model consists of the forward and backward processes. Formally, for an input sample $x_0$, the forward process tries to add noise:
\begin{equation}
\label{preliminaries2}
x_t = \gamma_tx_0 + \delta_t\epsilon, \quad \epsilon \sim \mathcal{N}(0,I),
\end{equation}
where $\gamma_t$ and $\delta_t$ are coefficients determined by the diffusion schedule, $\epsilon$ is the added Gaussian noise.

The reverse process learns to denoise and recover the data. The model's optimization objective is formulated as minimizing the reconstruction error:
\begin{equation}
\label{preliminaries3}
\mathcal{L} = \mathbb{E}_{\epsilon \sim \mathcal{N}(0,I)}[\|\epsilon - \epsilon_\theta(x_t, t, c)\|^2_2],
\end{equation}
where $x_t$ denotes the noisy sample at the $t$th step, $c$ denotes a condition vector for conditional generation. $\epsilon_\theta$ is the predicted error parameterized by $\theta$. To adapt the diffusion model for video processing, we make several modifications. 
First, we inflated the 2D convolutions to 3D convolutions~\cite{wu2023tune}. 
Then, following~\cite{wu2024towards}, we incorporated a parameter-efficient temporal attention module between the cross-attention and the feed-forward network.
These modifications enhance the coherence and consistency of the output video sequences.

Our M-GVR framework, illustrated in Fig.~\ref{sec:method}, is built on a pre-trained latent diffusion model~\cite{ldm} and comprises three main components: the Dual-stream Metadata Encoder (DME), the Prior-driven Mask Predictor (PMP), and the Post-Refinement Module (PRM). We introduce each of them in the following sections.
\subsection{Dual-stream Metadata Encoder}
To introduce metadata as conditions in the diffusion model, we need to map the metadata into the latent space.
For this purpose, we propose a dual-stream metadata encoder that processes motion vectors and frame types as distinct input conditions.
This encoder includes a Motion Vector Encoder and a Frame-type Encoder, each specifically designed to embed their respective metadata into the latent space.
\subsubsection{Motion Vector Encoder}
When the bitstream is corrupted, the motion vectors between frames are often directly affected, resulting in specific corruption patterns in the decoded video frames, such as block artifacts.
By incorporating motion vectors as a condition in the diffusion model, we enable it to better understand these corruption patterns and focus more precisely on the affected areas.
Additionally, motion vectors provide temporal information that guides consistent restoration across frames.

Our motion vector encoder consists of two 2D convolutional layers, each followed by a LeakyReLU activation function and an average pooling layer. We first extract motion vectors in standard H.264 format from compressed videos.
The raw motion vectors are two-dimensional vectors, where each row represents an individual motion vector, and each column contains motion information such as macroblock coordinates and motion displacements.
To capture block-wise movements between frames more explicitly, we map these vectors into the optical flow space.
Notably, $v$ includes concatenated optical flows from both preceding and subsequent frames. We input $v$ to the motion vector encoder to extract the spatial feature. A temporal transformer layer~\cite{wang2024videocomposer} is subsequently applied to capture temporal dependencies across frames, enhancing the model's ability to understand motion dynamics. 

\subsubsection{Frame-type Encoder}
Bitstream video generally uses the Group of Pictures (GOP) as its fundamental structure, comprising three types of frames: I-frames (Intra-coded picture), P-frames (Predicted picture), and B-frames (Bidirectional predicted picture).
Each frame type has unique dependencies on other frames, which helps in understanding how corruption in one frame might impact others.
For instance, an I-frame serves as an independent reference, and corruption in an I-frame may propagate to all dependent frames within the same GOP.
In contrast, a P-frame and B-frame rely on other frames, and their corruption follows a dynamic propagation pattern.
By integrating frame types, our model gains insight into inter-frame dependencies and how corruption is likely to spread, enabling a more effective and guided recovery process.

Specifically, we assign a one-hot vector to each frame type extracted from the bitstream and use a tokenizer to represent different frame types.
Next, two multi-layer perceptron layers are employed to map these tokens into feature embeddings.
The embeddings from the motion vector and the frame type branches are then combined via element-wise addition.
We project the embeddings in different formats, which are used in the denoising U-Net and the PMP, respectively, denoted as $r_m$ and $p_m$.

\subsection{Metadata as Diffusion Conditions}
After obtaining the metadata representation through the metadata encoder, we then integrate it into the denoising process. Specifically, we use a cross-attention mechanism, where the metadata representation serves as the key and value. The latent features act as queries. The output of the denoising process will be the intermediate recovery frame $\widetilde{y}$. Mathematically, we express the mapping $\Phi$ as:
\begin{equation}
\label{preliminaries4}
\widetilde{y}=\Phi(x,v,f),
\end{equation}
where $x$ is the input corrupted frame, $v$ is the motion vector and $f$ is the frame types. The training objective for optimization is:
\begin{equation}
\label{preliminaries5}
\mathcal{L}_d = \mathbb{E}_{\epsilon \sim \mathcal{N}(0,I), x, t}[\|\epsilon - \epsilon_\theta(x_l, t, c)\|^2_2],
\end{equation}where $x_l$ represents the latent of $x$ by a pre-trained auto-encoder, $c=(x, \tau_\theta(m, f))$ denotes the conditional inputs, $\tau_\theta(\cdot)$ embodies a dual-stream metadata encoder.
By doing so, our model can better address motion-induced corruption patterns and their propagation, improve the localization of corrupted regions and enable coherent recovery.
\subsection{Prior-driven Mask Predictor}
To achieve better recovery results, we aim to restore only the corrupted regions while preserving the integrity of intact areas.
To realize this, We propose a prior-driven mask predictor to generate pseudo masks that identify corrupted regions. This predictor leverages both diffusion and metadata features extracted earlier.
Specifically, we first interpolate the multi-scale attention outputs $att_d$ from the U-Net blocks to a uniform shape as the diffusion prior.
Then the diffusion prior concatenation with metadata prior $p_m$ and latent $x_l$ through a fusion module composed of five 3D convolutional layers with LeakyReLU activations. 
Finally, we use a pixel-unshuffle operation to upsample these features and generate the pseudo masks:
\begin{equation}
\label{preliminaries6}
\hat{m}=\Psi(x_l, p_m, att_d),
\end{equation}
where $\Psi$ denotes the prior-driven mask predictor and $\hat{m}$ is the generated pseudo mask. The training objective for optimization is defined as:
\begin{equation}
\label{preliminaries7}
\mathcal{L}_{m}=\mathrm{BCE}(m, \hat{m}),
\end{equation}
where $\hat{m}$ represents the predicted mask, and $m$ denotes the binarized residuals between corrupted frames and ground-truth frames. BCE denotes the binary cross entropy loss. This predictor focuses on accurately identifying corrupted regions, enabling targeted restoration without the need for labor-intensive mask annotations.

\subsection{Post-refinement Module}
Given the input frame $x$ and the intermediate result $\widetilde{y}$ generated by the U-Net, we use pseudo masks $\hat{m}$ to separate the intact region from $x$ and the recovered region from $\widetilde{y}$, combining them into the input $\widetilde{x}$ for the post-refinement module:
\begin{equation}
\label{preliminaries8}
\widetilde{x}=\hat{m}\odot \widetilde{y} + (1-\hat{m})\odot x,
\end{equation}
where $\odot$ denotes element-wise multiplication. The combined frames $\widetilde{x}$ are then processed through a 2D convolutional layer to extract shallow features.
Afterward, a cascade of residual swin transformer blocks~\cite{liang2021swinir} with identity connections is used to generate the final output. This process is expressed as:
\begin{equation}
\label{preliminaries9}
\hat{y}=\Theta(x,\widetilde{y},\hat{m}),
\end{equation}
where $\Theta$ represents the module for producing the final output $\hat{y}$. The training objective is a hybrid loss combines an $\mathrm{L1}$ loss $\mathcal{L}{l}$ and an adversarial loss using a temporal T-PatchGAN~\cite{chang2019free} discriminator:
\begin{equation}
\label{preliminaries10}
\begin{aligned}
\mathcal{L}_{a}=\mathbb{E}_{y\sim P_Y(y)}&[\mathrm{ReLU}(1-D(y))]\\ +& \mathbb{E}_{\hat{y}\sim P_{\hat{Y}}(y)}[\mathrm{ReLU}(1+D(\hat{y}))],
\end{aligned}
\end{equation}
where $D$ represents the discriminator, $P_Y(y)$ denotes represents the real data distribution, $P_{\hat{Y}}(y)$ denotes represents the predicted data distribution.

Eq.~\eqref{preliminaries8} represents a hard combination mechanism for generating $\widetilde{x}$. However, the predicted masks are not always perfect, this mechanism would lead to errors in the final results. Besides, while hard combination preserves intact regions, it may introduce artifacts along the boundary where regions are combined. We thus propose a post-refinement module (as shown in Eq.~\eqref{preliminaries9} and Eq.~\eqref{preliminaries10}) to enhance content consistency in the final results.
\section{Experiments}
\label{sec:Experiments}
\subsection{Settings and Implementations}

\noindent
{\bf Datasets:}
We utilize the BSCV~\cite{liu2024bitstream} dataset to train and evaluate our proposed M-GDM for bitstream-corrupted video recovery. 
This dataset provides a wide range of corruption patterns and severity levels, simulating real-world bitstream damages encountered during video transmission and storage.
Specifically, the BSCV dataset includes more than 28,000 bitstream-corrupted video clips (over 3,500,000 frames), derived from the most popular video inpainting datasets YouTube-VOS~\cite{yang20192nd} and DAVIS~\cite{pont20172017}. The video sequences are processed through H.264 video codec compression and segments are randomly removed to simulate packet loss and storage-related errors. The resulting corruption patterns include block artifacts, color distortions, duplication, misalignment, texture loss, and trailing artifacts. These features make BSCV distinct from existing inpainting datasets, which rely on simulated, manually-designed binary error masks. We use 3471 video sequences with the corruption parameter of (1/16, 0.4, 4096) to train our network.
The validation will be performed under two different subsets: YouTube-VOS and DAVIS, which consist of 508 and 50 video clips, respectively.

\noindent
{\bf Training Details:}
We initialize the U-Net weights from \cite{wu2024towards}.
We first fine-tune the U-Net on YouTube-VOS for $100k$ iterations, using the Adam optimizer with a learning rate of 1e-5.
Since there are no text prompts in YouTube-VOS, we use the prompt ``null'' during fine-tuning.
Then, we train another $100k$ iterations for dual-stream metadata encoder and prior-driven mask predictor, where the learning rates are 1e-6, 1e-4, and 1e-4 for U-Net, dual-stream metadata decoder, and prior-driven mask predictor, respectively. 
The loss weights for this phase are set to $\lambda_1$=1, $\lambda_2$=1e-3 for Equ.~\eqref{preliminaries5} and Equ.~\eqref{preliminaries7}.
Finally, we freeze the trained network and proceed to train the post-refinement module for $50k$ iterations, with a learning rate of 1e-5.
Each video sequence used in both training and evaluation consists of 16 frames, with each frame resized to 448 $\times$ 256.
All training stages are conducted on 16 NVIDIA H20 GPUs with a total batch size of 64. 

\noindent
{\bf Metrics:}
We use a range of metrics to evaluate both the frame quality and temporal coherence of the generated results. 
For frame quality, we employ PSNR and SSIM to assess the similarity between predicted results and ground truth. We use VFID~\cite{kim2019deep} and LPIPS~\cite{zhang2018unreasonable} to measure the perceptual similarities, temporal consistency, and smoothness.

\noindent
{\bf Baselines:}
We compare our M-GDM with $\mathrm{E}^2$FGVI~\cite{li2022towards}, ProPainter~\cite{zhou2023propainter} and BSCVR~\cite{liu2024bitstream}. $\mathrm{E}^2$FGVI is an end-to-end flow-guided framework with masked frames. ProPainter uses a mask-guided sparse video transformer for image and feature propagation, while BSCVR further incorporates corrupted content to enhance features of partially damaged areas. Since these methods require additional masks to indicate the corrupted areas, we use SAM2~\cite{ravi2024sam2}, which has been finetuned for better segmentation performance, to generate pseudo masks for each frame. For a fair comparison, these predicted masks also undergo dilation operations before sending to the inpainting models following~\cite{liu2024bitstream}.
\begin{figure*}[t]
  \centering
   \includegraphics[width=\linewidth]{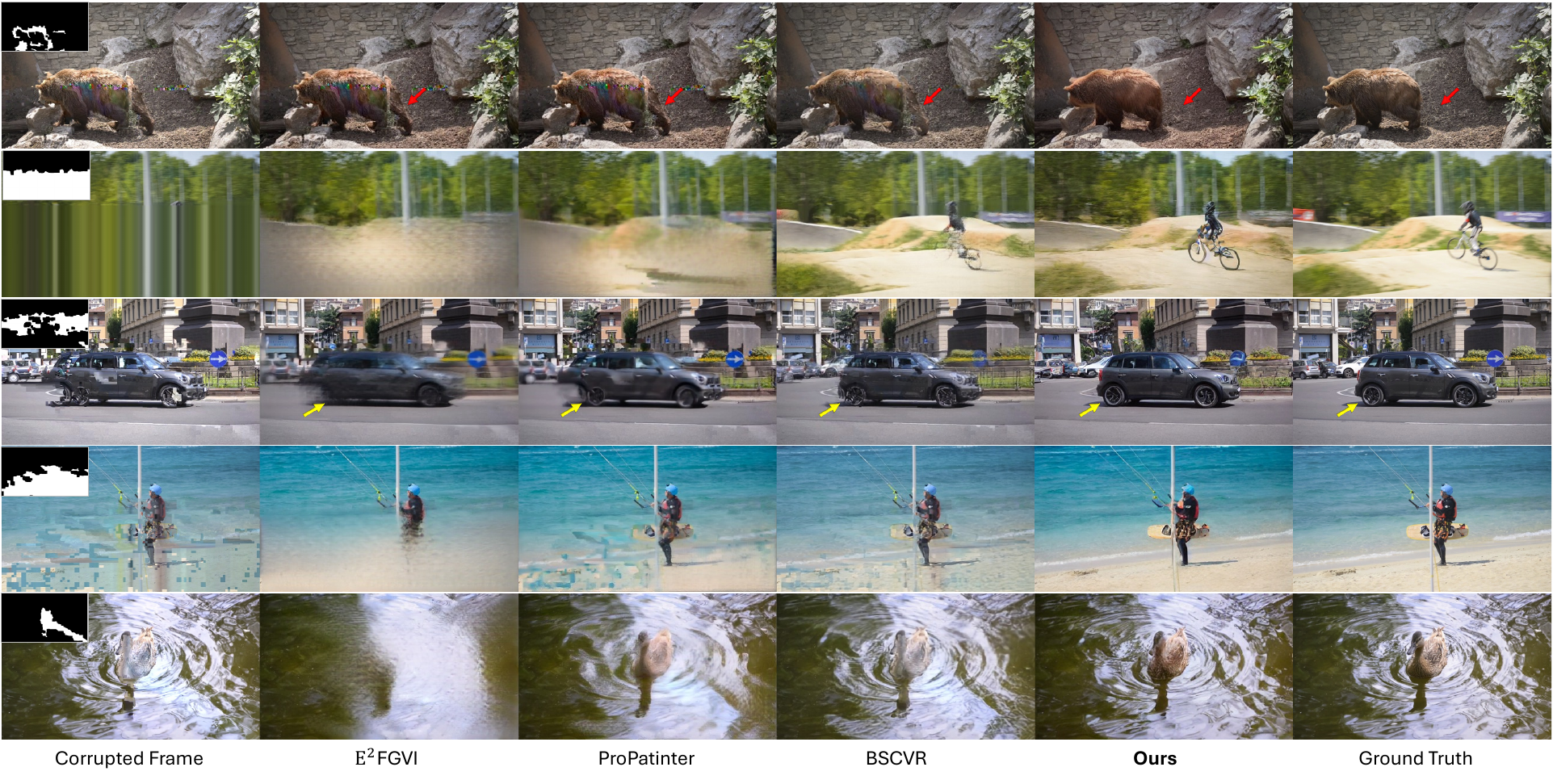}
   \vspace{-1.5em}
    \caption{\textbf{Qualitative Comparison on the YouTube-VOS Dataset.} From left to right, we present the corrupted frames (with the ground-truth masks shown in the top-left corner), the results of $\mathrm{E}^2$FGVI, ProPainter, BSCVR, our proposed M-GDM, and the ground-truth. Our approach consistently produces results that are visually more appealing.}
   \label{fig:exp_1}
\end{figure*}

\subsection{Quantitative Comparison}
We first present the bitstream-corrupted video recovery results on the YouTube-VOS dataset (see Table~\ref{tab:1}). 
Our proposed M-GDM consistently achieves notable improvements over baselines including $\mathrm{E}^2$FGVI, ProPainter, and BSCVR across all evaluation metrics. 
Specifically, our M-GDM achieves the highest PSNR of 28.21, outperforming the closest competitor, BSCVR, by a margin of 0.53. 
Moreover, M-GDM achieves a lower LPIPS value of 0.0333, indicating superior perceptual quality with reduced perceptual deviation from the ground truth frames.
We also report our recovery results on the DAVIS dataset in Table~\ref{tab:2}. 
The consistent improvements over the baselines validate the effectiveness of M-GDM. 
Specifically, our M-GDM achieves the highest PSNR value of 26.05, exceeding BSCVR by 1.55, which indicates superior reconstruction quality.
Additionally, our model obtains the lowest VFID value of 0.1621, reflecting enhanced temporal consistency and stability in the reconstructed video sequences.
Overall, these results highlight M-GDM's effectiveness in addressing bitstream-corrupted video recovery, achieving high reconstruction quality, structural similarity, and temporal consistency.

\begin{table}[t]
    \centering
    \caption{\textbf{Quantitative Comparison on the YouTube-VOS Dataset.} We compare with the results of state-of-the-art video recovery methods on the YouTube-VOS dataset. Our proposed M-GDM achieves the best performance across all the metrics.}
    \vspace{-1em}
    \renewcommand{\arraystretch}{1.15}
    \resizebox{\linewidth}{!}{
        \begin{tabular}{l|cccc} 
            \toprule
             Method &  PSNR$\uparrow$ & SSIM$\uparrow$ & LPIPS$\downarrow$  &  VFID$\downarrow$ \\
            \hline \hline 
            $\mathrm{E}^2$FGVI~\cite{li2022towards} & 26.26 & 0.8774 & 0.0547 & 0.0783 \\
             ProPainter~\cite{zhou2023propainter}  & 27.01 & 0.9024 & 0.0439 & 0.0681 \\        
             BSCVR~\cite{liu2024bitstream} &  27.68 & 0.9173 & 0.0359 & 0.0662 \\
            \hline
            M-GDM (Ours) & \textbf{28.21} & \textbf{0.9202} & \textbf{0.0333} & \textbf{0.0661} \\  
            \bottomrule
        \end{tabular}
        \label{tab:1}
    }
\end{table}

\begin{table}[t]
    \centering
    \caption{\textbf{Quantitative Comparison on the DAVIS Dataset.} We report the performance of different video recovery methods on the DAVIS dataset. Our M-GDM achieves the highest values across all the metrics.}
    \vspace{-1em}
    \renewcommand{\arraystretch}{1.15}
    \resizebox{\linewidth}{!}{
        \begin{tabular}{l|cccc} 
            \toprule
             Method & PSNR$\uparrow$ & SSIM$\uparrow$ & LPIPS$\downarrow$  &  VFID$\downarrow$ \\
            \hline \hline 
            $\mathrm{E}^2$FGVI~\cite{li2022towards} & 23.83 & 0.8272 & 0.0569 & 0.2401  \\  
            ProPainter~\cite{zhou2023propainter} & 24.13 & 0.8436 & 0.0510 & 0.2159 \\        
            BSCVR~\cite{liu2024bitstream} & 24.50 & 0.8613 & 0.0433 & 0.1864 \\   
            \hline
            M-GDM (Ours) & \textbf{26.05} & \textbf{0.8822} & \textbf{0.0340} & \textbf{0.1621} \\  
            \bottomrule
        \end{tabular}
        \label{tab:2}
    }
\end{table}

\begin{figure*}[t]
  \centering
   \includegraphics[width=\linewidth]{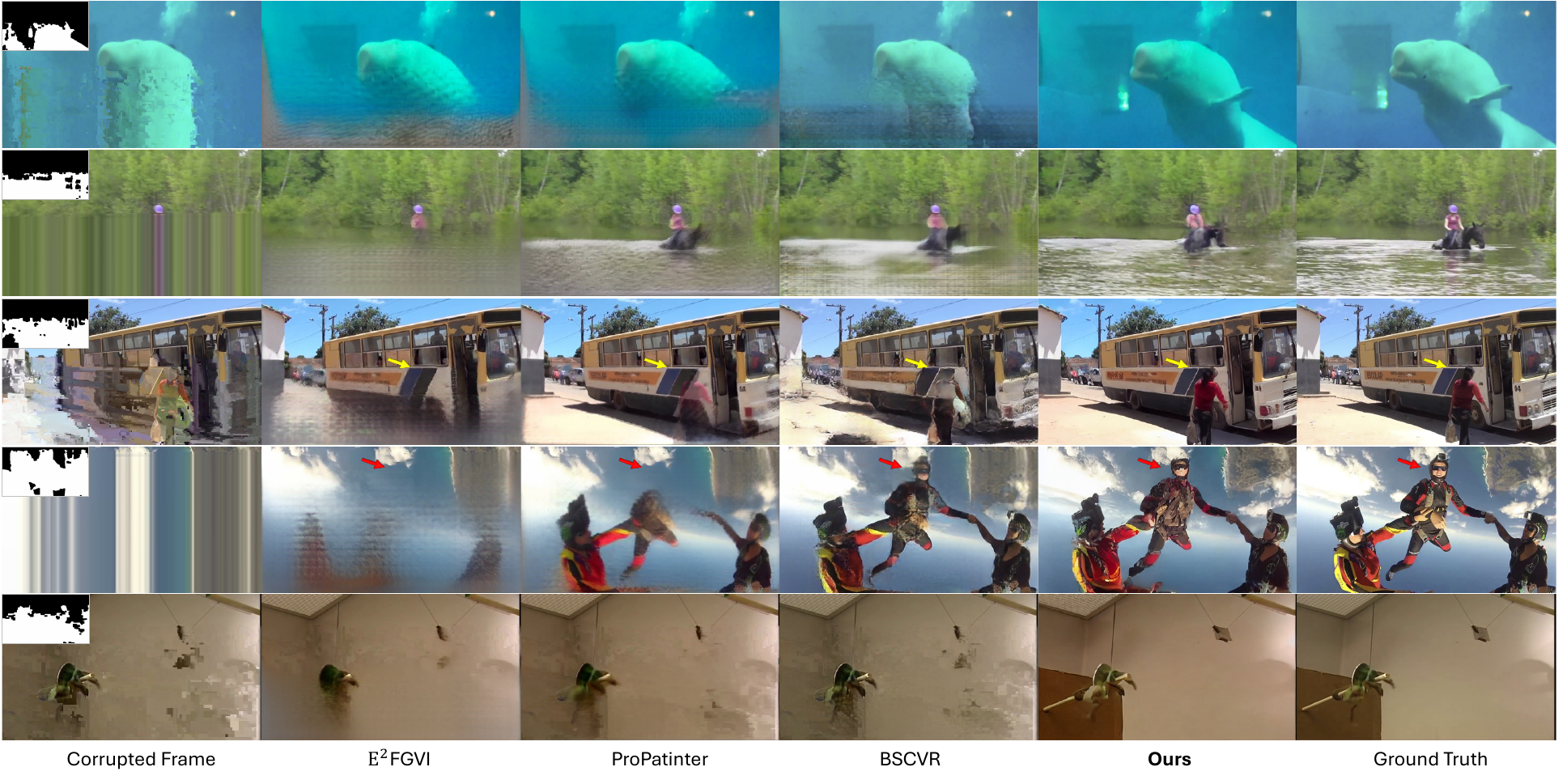}
    \caption{\textbf{Qualitative Comparison on the DAVIS Dataset.} We showcase the qualitative comparison among different video recovery methods on DAVIS. The artifacts caused by bitstream corruption have been significantly reduced by our method while the other methods still suffer obvious artifacts.
    \vspace{-1em}
    }
   \label{fig:exp_2}
\end{figure*}
\begin{table}[t]
    \centering
    \caption{\textbf{Ablation Studies of the Proposed Components.} DME represents the Dual-stream Metadata Encoder, PMP is the Prior-driven Mask Predictor, and PRM indicates the Post-Refinement Module. Starting from the fine-tuned temporal U-Net (a), we progressively add each component for evaluation.}
    \vspace{-1em}
    \renewcommand{\arraystretch}{1.15}
    \resizebox{\linewidth}{!}{
    \begin{tabular}{c|ccc|ccc} 
        \toprule
        Exp. & DME & PMP & PRM & PSNR$\uparrow$ &SSIM$\uparrow$& LPIPS$\downarrow$ \\   
        \hline \hline 
        (a) & & &  & 21.56 & 0.8054 & 0.0565\\
        (b) & \checkmark &  &  &22.27 & 0.8122 & 0.0538  \\
        (c) & \checkmark & \checkmark & & 25.84 & 0.8784 & \textbf{0.0328}\\
        (d) & \checkmark & \checkmark & \checkmark & \textbf{26.05} & \textbf{0.8822} & 0.0340  \\
        \bottomrule
    \end{tabular}
    \label{tab:3}
    }
\end{table}
\subsection{Qualitative Comparison}
We present qualitative analysis by comparing M-GDM with the vanilla baselines in Fig.~\ref{fig:exp_1} and Fig.~\ref{fig:exp_2}.
As shown in the figures, our method consistently produces results that are visually closer to the ground truth, with significantly fewer artifacts than the baselines.
In challenging scenarios with complex textures or significant motion (e.g., water ripples or moving vehicles), M-GDM excels in preserving fine details and ensuring smoother transitions.
For example, in the bear and water ripple scenes, our method produces clearer, artifact-free visual reconstructions, while the baselines either blur important details or introduce unnatural distortions. Moreover, in the bus and skydiving scenes, M-GDM maintains structural integrity and reduces boundary artifacts, evidenced by sharper edges and more consistent color distribution.
These qualitative results underscore M-GDM’s capability to maintain high visual fidelity and temporal consistency across frames, demonstrating its effectiveness in addressing the bitstream-corrupted video recovery problem compared to existing methods. More visual results and recovered final videos are provided in the supplementary materials.

\subsection{Ablation Studies}
We conduct an ablation study in Table~\ref{tab:3} to investigate the contribution of each component in our framework. Specifically, we start with a fine-tuned temporal U-Net and progressively add the dual-stream metadata encoder (DME), prior-driven mask predictor (PMP), and post-refinement module (PRM).

\noindent
{\bf Effect of metadata guidance:}
By comparing experiments (a) and (b) in Table~\ref{tab:3}, we can observe the effect of metadata guidance on video recovery performance. Using metadata to guide the diffusion model significantly improves PSNR (from 21.56 to 22.27) and LPIPS (from 0.565 to 0.538).
This demonstrates that leveraging metadata allows the model to better understand the corrupted regions, resulting in enhanced reconstruction quality.

\noindent
{\bf Benefits of introducing mask predictor:}
Experiment (c), which includes the PMP, shows substantial improvements over experiment (b).
The PSNR increases from 22.27 to 25.84, and the LPIPS significantly decreases from 0.0538 to 0.0328.
The benefits of PMP can be attributed to two main factors: (1) the mask supervision helps the diffusion model to focus more on the corrupted regions, and (2) with the generated masks, the hard combination mechanism ensures the preservation of intact areas, minimizing unnecessary modifications.

\noindent
{\bf Effectiveness of refinement module:}
Experiment (d), which integrates the PRM, achieves a PSNR of 26.05 and an SSIM of 0.8822.
The LPIPS value remains competitive at 0.0340, with slight fluctuations. 
This is likely due to the refinement process, which aims to eliminate boundary artifacts but may also smooth high-frequency details. Nonetheless, the refinement module effectively enhances the overall content consistency between recovered and intact regions, leading to a more coherent final output.
\section{Conclusion}
In this paper, we introduced a novel framework, Metadata-Guided Diffusion Model (M-GDM), specifically for blind bitstream-corrupted video recovery. Unlike traditional methods that heavily rely on predefined masks, M-GDM leverages inherent video metadata to automatically identify and recover corrupted regions. Our approach eliminates the need of annotated masks as input and thus is more practical. Compared to previous methods, our M-GDM achieves superior visual recovery performance in terms of both frame quality and temporal consistency. 
Though our recovered results are visually promising, we also observed there exists slight color deviation from original corrupted videos. This is probably due to the adopted video stable diffusion model that was originally designed for video generation rather than recovery. We believe that by designing a network more suitable for video recovery, the color deviation phenomenon can be effectively alleviated. 
Overall, M-GDM sets a new standard for blind bitstream-corrupted video restoration and sheds light on video metadata-guided recovery tasks.

\subsection*{Acknowledgments} 
This research is funded in part by ARC-Discovery grant (DP220100800 to XY) and ARC-DECRA grant (DE230100477 to XY). We thank all anonymous reviewers and ACs for their constructive suggestions.
\label{sec:Conclusion}

{
    \small
    \bibliographystyle{ieeenat_fullname}
    \bibliography{main}
}
\end{document}